\documentclass[10pt,twocolumn,letterpaper]{article}

\usepackage{iccv}
\usepackage{times}
\usepackage{epsfig}
\usepackage{graphicx}
\usepackage{amsmath}
\usepackage{amssymb}

\usepackage[accsupp]{axessibility}

\usepackage{booktabs}
\usepackage{multirow}
\usepackage{gensymb}
\usepackage{colortbl}
\usepackage{xcolor}

\newtheorem{lemma}{Lemma}

\usepackage{capt-of,etoolbox}
\usepackage{caption}

\usepackage{color}
\definecolor{mycitecolor}{rgb}{0, 0.4, 0.7}
\usepackage[pagebackref=true,breaklinks=true,letterpaper=true,colorlinks,bookmarks=true,citecolor=mycitecolor]{hyperref}

\iccvfinalcopy 

\def\iccvPaperID{9206} 

\ificcvfinal\fi

\begin{document}

\title{Density-invariant Features for Distant Point Cloud Registration}

\author{Quan Liu\ \ \ \ \ \ \ \ \ \ Hongzi Zhu\thanks{Corresponding author}\ \ \ \ \ \ \ \ \ \ Yunsong Zhou\\
Shanghai Jiao Tong University\\
{\tt\small \{liuquan2017, hongzi, zhouyunsong\}@sjtu.edu.cn}
\and
Hongyang Li\\
Shanghai AI Lab\\
{\tt\small hy@opendrivelab.com}
\and
Shan Chang\\
Donghua University\\
{\tt\small changshan@dhu.edu.cn}
\and
Minyi Guo\\
Shanghai Jiao Tong University\\
{\tt\small guo-my@cs.sjtu.edu.cn}
}
\twocolumn[{%
\renewcommand\twocolumn[1][]{#1}%
\maketitle
\begin{center}
    \centering
    \captionsetup{type=figure}
    \vspace{-0.5cm}
    \includegraphics[width=\linewidth]{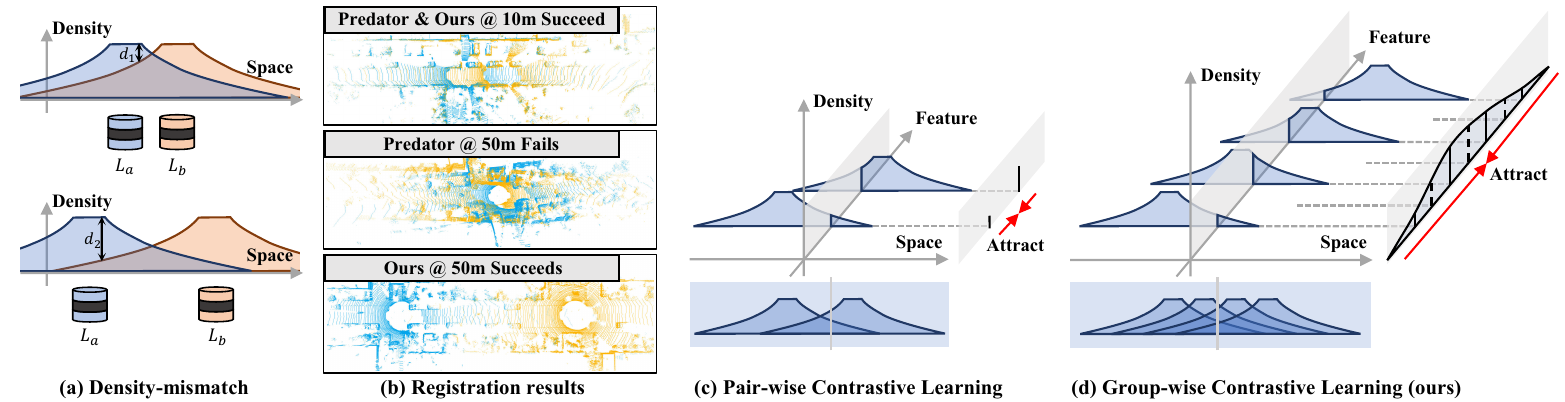}
    \vspace{-0.6cm}
    \captionof{figure}{\textbf{Motivation.} (a) A schematic diagram of the density $d$ of point clouds produced by two LiDARs $L_a$ and $L_b$. Their densities diverge when two LiDARs drift from short-range to long-range scenario. (b) State-of-the-art method Predator~\cite{huang2021predator} fails to register distant point clouds (middle) while our method succeeds (bottom). (c) Traditional Pair-wise Contrastive Learning takes a pair of positive samples (marked with the gray plane) from two point clouds, and pull positive features together, which suffer from negative density correlation given a fixed pair of distant point clouds. (d) Our Group-wise Contrastive Learning takes multiple positive samples (marked with the gray plane) from a point cloud series, and contract the feature distance between all positive examples.}
    \label{fig:difficulty}
\end{center}%
}]
\ificcvfinal\thispagestyle{empty}\fi

\let\thefootnote\relax
\footnotetext{* Corresponding author}

\begin{abstract}
Registration of distant outdoor LiDAR point clouds is crucial to extending the 3D vision of collaborative autonomous vehicles, and yet is challenging due to small overlapping area and a huge disparity between observed point densities.
In this paper, we propose Group-wise Contrastive Learning (GCL) scheme to extract density-invariant geometric features to register distant outdoor LiDAR point clouds. We mark through theoretical analysis and experiments that, contrastive positives should be independent and identically distributed (i.i.d.), in order to train density-invariant feature extractors. We propose upon the conclusion a simple yet effective training scheme to force the feature of multiple point clouds in the same spatial location (referred to as positive groups) to be similar, which naturally avoids the sampling bias introduced by a pair of point clouds to conform with the i.i.d.~principle. The resulting fully-convolutional feature extractor is more powerful and density-invariant than state-of-the-art methods, improving the registration recall of distant scenarios on KITTI and nuScenes benchmarks by 40.9\% and 26.9\%, respectively. Code is available at https://github.com/liuQuan98/GCL.
\end{abstract}
\vspace{-0.5cm}
\section{Introduction}
Point cloud registration is the cornerstone technique for various computer vision tasks such as SLAM \cite{montemerlo2002fastslam,mur2017orb}, scene flow estimation \cite{liu2022camliflow,liu2019flownet3d}, and early/late fusion in 3D scene understanding \cite{zhang2021emp,yu2022dair,ku2018joint,zhu2022vpfnet}.
Due to the complex scenarios and the large scale of outdoor point clouds, point cloud registration in driving scenarios is a more challenging and rewarding task than indoor scenes, which can help expand the perceptual field of collaborative vehicles for enhancing the driving safety.
In such scenarios, point cloud registration should be accurate enough even when dealing with \emph{extremely low-overlap} point clouds (\emph{e.g.}, two LiDARs of interest may be over 50 meters apart) to secure downstream tasks such as object detection \cite{zhu2022vpfnet, ku2018joint}, segmentation \cite{xu2020squeezesegv3}, and tracking \cite{wu20213d,simon2019complexer}.




As depicted in Figure \ref{fig:difficulty}(a), as a pair of overlapping point clouds drift apart, they scan
the same location with increasingly different densities because point cloud local density reduces quadratically with the distance from the LiDAR. This is referred to as the \textit{density-mismatch} problem exclusively found on distant outdoor point clouds. As generally discovered by literature~\cite{wu2019pointconv,te2018rgcnn,yew20183dfeat,huang2021predator,qin2022geometric}, deep learning features are sensitive to point density so that density-mismatch results in low feature similarity
and harms registration performance.


In recent years, there has been a boom in learning-based outdoor point cloud registration methods~\cite{deng2018ppfnet,gojcic2019perfect,poiesi2021distinctive,ao2021spinnet,yew20183dfeat,choy2019fully,bai2020d3feat,huang2021predator,yew2022regtr}, all of which train feature extractors based on a pair of point clouds, referred to as the pair-wise contrastive learning (PCL) technique \cite{deng2018ppfnet}. 
As stated by Arora \etal \cite{arora2019theoretical}, a prerequisite for PCL network convergence is that data samples in a positive pair, which represents scans of a location made by two LiDARs, have to be independently and identically distributed (\emph{i.i.d.}). However, severe density-mismatch leads to violation of the \emph{i.i.d.}~principle. More specifically, given a pair of distant point clouds $S,T$, it is highly likely that a high-density location in point cloud $S$ corresponds to a low-density location in point cloud $T$ and vise versa, \emph{i.e.}, their densities are correlated, as depicted in Figure \ref{fig:difficulty}(c). Despite adopting several density-related techniques such as voxelization \cite{gojcic2019perfect, lu2021hregnet, yu2021cofinet, qin2022geometric, yew20183dfeat, choy2019fully, huang2021predator, yew2022regtr} or density-adaptive calculation \cite{bai2020d3feat,wu2019pointconv}, existing methods still fail when handling distant outdoor point clouds, as depicted in Figure \ref{fig:difficulty} (b), where state-of-the-art method Predator \cite{huang2021predator} fails to register two point clouds of 50 meters apart.

Based on the analysis above, we remark that \emph{i.i.d.}~positive features \emph{w.r.t.}~density are essential for training \textit{density-invatiant} feature extractors to solve the distant point cloud registration problem.

In this paper, we propose \emph{group-wise contrastive learning} (GCL) scheme for \emph{density-invariant} feature extraction in order to register distant point clouds. The core idea of GCL is to utilize groups of highly-overlapped point clouds, continuously collected by moving vehicles of public datasets, to break the density correlation between positive examples. As illustrated in Figure \ref{fig:difficulty}(d), GCL aligns such point clouds, and collects all point correspondences and their features at one location, referred to as a \emph{positive group}. A large enough positive group can better approximate the underlying feature distribution. Furthermore, given a positive group and a specific positive sample in the group, the density of its possible correspondence is unknown since we do not know which point cloud the other sample belongs to. As a result, GCL positive samples are approximately \emph{i.i.d.}~and can be used for designing compelling group-wise loss to train the density-invariant feature extractor.


How to engage the extractor to derive consistent features over inconsistent densities in positive groups is non-trivial.
One straightforward solution is to minimize the variance of all features, which is not sufficient to set the optimal convergence target. In contrast, we additionally ask the mean of a positive group to be close to its \textit{finest feature}, which is defined as the feature derived from the point with the highest point density in the group. By doing this, we force features extracted with low point densities to be similar with the most descriptive one in a positive group, facilitating features in a positive group to converge towards a better consensus.

We implement GCL on both sparse voxel convolution and KPConv, and conduct extensive experiments on KITTI \cite{Geiger2012CVPR} and nuScenes \cite{Caesar_2020_CVPR}.
Results demonstrate that GCL can achieve above 40.9\% and 26.9\% registration recall gains over SOTA point cloud registration methods when handling far point cloud pairs in KITTI and nuScenes, respectively, without performance loss on near point cloud registration benchmarks.
In addition, GCL is lightweight, making it preferable for online distant point cloud registration on smart vehicles.

We highlight our main contributions as follows:
\begin{itemize}
    \item We theoretically analyze the difficulty of registering distant point clouds, and mark that constructing more  \emph{i.i.d.}~positive samples is the key to train a density-invariant feature extractor for this challenging task.
    \item
    We propose an effective density-invariant feature extraction scheme based on group-wise contrastive learning, where \emph{i.i.d.}~positive groups are neatly constructed and new contrastive learning loss are particularly designed.
    \item We conduct extensive trace-driven experiments on KITTI and nuScenes. Results demonstrate superior density-invariance along with +40.9\% and +26.9\% registration recall
    improvements on distant scenarios in KITTI and nuScenes, respectively.
\end{itemize}

\vspace{-0.2cm}
\section{Related Work}



\subsection{Deep Point Cloud Registration}
Recent registration pipelines often adopt a learning-based 3D backbone with contrastive loss to extract local geometry for feature matching. This is the main focus in this work.

\begin{figure*}[t]
  \centering
  \includegraphics[width=1\linewidth]{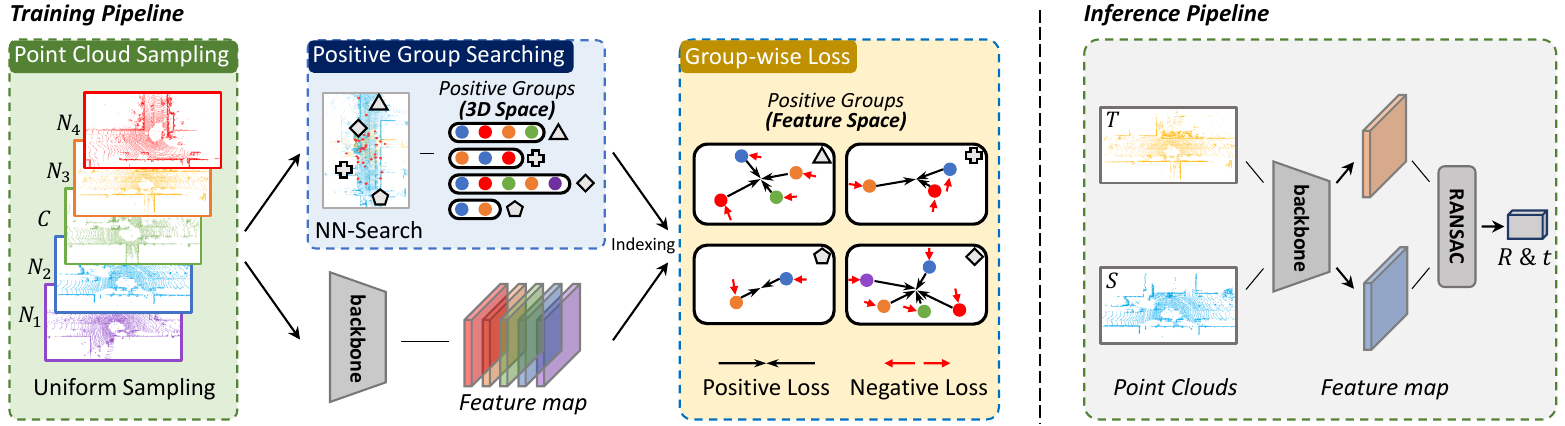}
  \caption{\textbf{Overview} for the proposed Group-Wise Contrastive Learning (GCL). The training pipeline of GCL composes of three stages: (1) Point Cloud Sampling, where multiple neighboring point clouds are uniformly sampled around a central point cloud; (2) Positive Group Searching, which collects all correspondences on the same spatial location to form a positive group, then collects their features from the feature map; (3) Group-wise Loss, where contrastive loss is applied on positive groups to forge density-invariant features. During inference, features are extracted for a pair of distant point clouds and then fed into RANSAC \cite{fischler1981random} to calculate the relative transformation.
  }
  \label{fig:overview}
  \vspace{-0.4cm}
\end{figure*}

\vspace{-0.2cm}
\paragraph{Patch-based features.} Patch based features \cite{zeng20173dmatch,deng2018ppfnet,gojcic2019perfect,poiesi2021distinctive,ao2021spinnet} generally follow the pioneering 3DMatch \cite{zeng20173dmatch} to extract deep features on pre-selected local patches, and apply contrastive loss on patch embeddings. PPF-Net \cite{deng2018ppfnet} further introduces the PointNet backbone \cite{qi2017pointnet}, while PerfectMatch \cite{gojcic2019perfect} improves feature robustness with smoothed density value. DIP \cite{poiesi2021distinctive} incorporates a patch reconstruction step. SpinNet \cite{ao2021spinnet} uses spherical convolution to achieve SO(2) equivalence. However, patch-based methods are usually slow due to repeated computation even when patches largely overlap each other, hindering their application in real-time scenarios such as self-driving.

\paragraph{Fully convolutional features.} Following FCGF \cite{choy2019fully}, fully convolutional methods \cite{choy2019fully,bai2020d3feat,huang2021predator} extract dense features for the whole point cloud in one forward pass and apply contrastive loss to points instead of patches. These methods achieve both state-of-the-art performance and low inference time. D3Feat \cite{bai2020d3feat} grants the KPConv \cite{thomas2019kpconv} extractor the ability of a key-point detector.
Predator \cite{huang2021predator} further improves the low-overlap scenario with an overlap attention module in the bottleneck. Despite their promising performance in indoor or close LiDAR point clouds, their negligence of density variance leads to degraded performance on distant outdoor point clouds, which this paper fixates on improving.


\subsection{Methods against Density Variation}
The conventional techniques aimed at addressing density variation have been extensively used in point cloud processing. However, they are insufficient to solve the distant point cloud registration problem. Voxelization slightly eases density variation and has been adopted by SOTA methods \cite{gojcic2019perfect,lu2021hregnet,yu2021cofinet,qin2022geometric,yew20183dfeat,choy2019fully,huang2021predator,yew2022regtr}, but has limited effect according to Fig. \ref{fig:difficulty}(b). Sampling methods such as Farthest Point Sampling \cite{eldar1997farthest, qi2017pointnet, aoki2019pointnetlk, sarode2019pcrnet} achieve uniform density by aggressively dropping points, undermining feature descriptiveness. Density estimation methods, \emph{e.g.}, distance-based \cite{lawin2018density}, KDE \cite{turlach1993bandwidth, wu2019pointconv} or SDV \cite{gojcic2019perfect}, allow density-adaptive feature calculation, but also requires sufficient input sampling to work properly, which is parallel with our work.

\subsection{Contrastive Learning}
Contrastive Learning, also known as Deep Metric Learning, was first introduced to effectively extract dense visual representations \cite{hadsell2006dimensionality,schroff2015facenet,kumar2016learning,hermans2017defense,wu2018unsupervised,chen2020simple,he2020momentum,wang2020cross}, then extended to process audio \cite{wu2022wav2clip} or texts \cite{gao2021simcse,jia2021scaling}. A noticeable trend is that sampling more and harder negative samples will improve feature quality due to elevated stability and informativeness \cite{oh2016deep,hermans2017defense,wu2018unsupervised,deng2018ppfnet,he2020momentum,wang2020cross}, but improvements on positive samples are scarce.
Our work is the first to examine density variation of 3-dimensional point clouds and apply loss on positives with multiple different densities, which a topic that has not been well studied yet.

\section{Method}
The GCL method overview is depicted in Figure \ref{fig:overview}, which adopts a feature-based architecture~\cite{choy2019fully,bai2020d3feat,huang2021predator}. During training, GCL composes of three stages: (1) Point Cloud Sampling from neighboring views that are uniformly distributed on the road; (2) Positive Group Searching which finds positive groups containing observations of the same spot from various perspectives and distances, and collects their corresponding features; (3) Group-wise Loss that applies advanced constraints on positive groups instead of the positive pair loss adopted in PCL. 

During inference, features are extracted on two point clouds $S, T$ before being fed into a robust estimator such as RANSAC \cite{fischler1981random} to recover the relative transformation.

\subsection{Preliminary}
\paragraph{Point cloud registration.} Given two distant point clouds $S=\{p_S^i\in\mathbb{R}^3|i=1,2,...,n\},  T=\{p_T^j\in\mathbb{R}^3|j=1,2,...,m\}$ with partial overlap, the point cloud registration problem is to uncover the optimal transformation $R\in SO(3), t\in \mathbb{R}^3$ that resembles the spatial displacements between both LiDARs.

\paragraph{Pair-wise Contrastive Learning in registration.} Contrastive learning is a pairwise optimization based training paradigm, where data in the same class (positives) are encouraged to have similar features, and data in different classes (negatives) are encouraged to have distinct features. In the context of point cloud registration, positives refer to points in the same spatial location (\emph{i.e.}, correspondences) and negatives refer to points in different locations. In practice, PCL methods \cite{deng2018ppfnet,gojcic2019perfect,poiesi2021distinctive,ao2021spinnet,yew20183dfeat,choy2019fully,bai2020d3feat,huang2021predator,yew2022regtr} usually find point correspondences on two point clouds, and apply contrastive loss on the corresponding features for supervision. They usually have a RANSAC-based inference pipeline as the one depicted in Figure \ref{fig:overview}. 

\subsection{Analysis}
\label{sec:analysis}


\begin{figure}[t]
  \centering
  \includegraphics[width=0.7\linewidth]{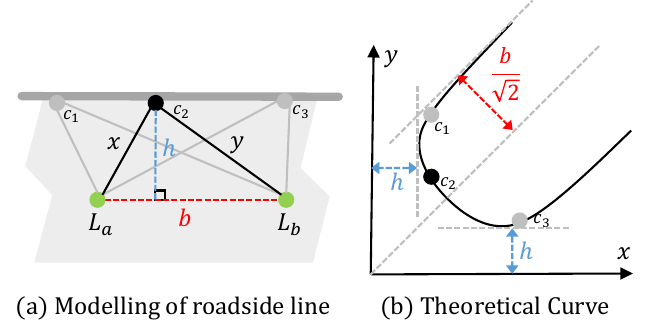}
  \caption{\textbf{Our hypothesis.} (a) Looking from bird's eye view, PCL correspondences $c$ cluster on roadside lines, and (b) they form a `U' shaped curve when plotting the distance from a correspondence to both LiDARs $L_a, L_b$ as the $x$-$y$ coordinate. For example, three spatially collinear correspondences $c_1,c_2,c_3$ are bent with the `U' curve when using $x$ and $y$ as their coordinates.
  }
  \vspace{-0.4cm}
  \label{fig:hypothesis}
\end{figure}

\paragraph{\textit{I.i.d.}~positive features.} It is a general basic assumption that positives should be sampled in an \textit{i.i.d.}~manner \cite{arora2019theoretical}, so that the positive features of PCL can converge together to achieve density invariance. Given a specific location in world coordinates, we denote its feature distribution $D$ as containing all possible features for this specific location observed from all densities and angles. We denote a positive pair as $(p_S^i,p_T^j)\in C$, and their features as $f_S^i, f_T^j\sim D$ from point cloud $S$ and $T$, respectively. Note that $S$ and $T$ are also variables. $C$ denotes all pairs of positive correspondences in this spatial location. The positive loss $L_{pos}$
on this location is formulated in Equation \ref{eq:pos}, where parameter $r\in[1,+\infty)$ represents vector norm, and $M\in \mathbb{R}^+$ is a tolerance margin.
\label{sec:iid}

\vspace{-0.3cm}
\begin{equation}
    L_{pos}=\frac{1}{|C|}\sum_{(p_S^i,p_T^j)\in C}{max(||f_S^i-f_T^j||_r-M, 0)}
    \label{eq:pos}
    \vspace{-0.3cm}
\end{equation}

\begin{lemma}
\label{lemma:iid}
\textit{If $f_S^i,f_T^j\sim D$ are i.i.d., then $\exists \hat{f}$, so that minimizing $L_{pos}$ converges in probability to encouraging all features $f\sim D$ to converge towards the same location $\hat{f}$ in feature space; Otherwise, non-i.i.d.~sampling of $f_S^i,f_T^j$ will encourage all features to converge towards different locations in feature space.}\\
\end{lemma}
\vspace{-0.5cm}

Detailed proof is provided in Appendix \ref{proof:iid}. Lemma \ref{lemma:iid}
is the fundamental reason why PCL cannot handle distant point clouds. Besides, density-invariance can be achieved through features of all densities converging to the same location. This hints that positive features must be independent in order to train a density-invariant feature extractor.

\vspace{-0.3cm}

\paragraph{General impact of density-mismatch on PCL.}
Ideally, positive features should be independently sampled from all over the space. However, this is not true for PCL due to the sampling bias introduced by distant point clouds. PCL samples positives in the limited overlap of a pair of point clouds, trying to capture everything in the overlap while ignoring others out of the overlap. We will show that the spatially restricted sampling strategy in turn breaks the \textit{i.i.d.}~assumption of positive features.
\label{sec:density_correlation}
\vspace{-0.3cm}

\begin{figure}[t]
  \centering
  \includegraphics[width=1\linewidth]{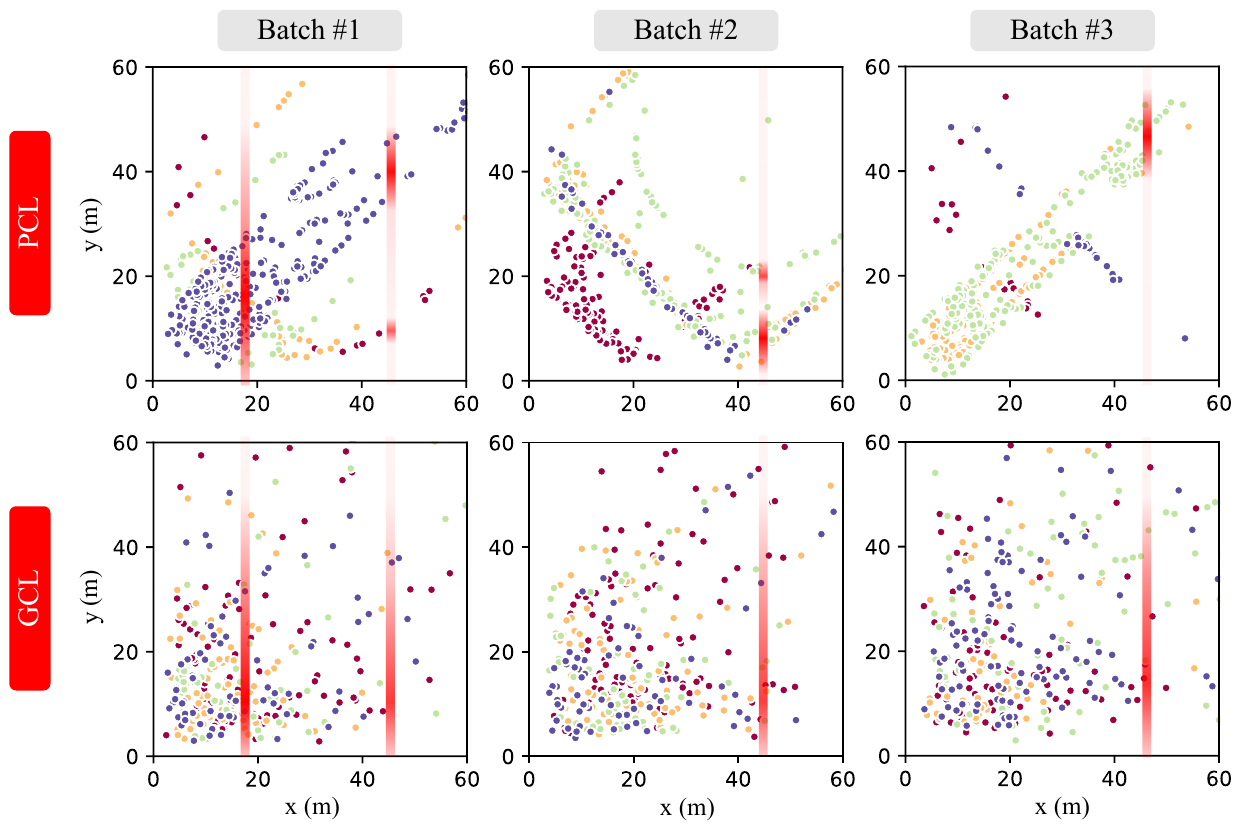}
  \caption{\textbf{Visual validation} of our hypothesis in Figure \ref{fig:hypothesis}. We plot the curves of real correspondences in PCL (top row) and GCL (bottom row) with $batch\_size=4$, where the $x$ and $y$ coordinates of a dot denotes the distance from a correspondence to both LiDARs. Points are colored according to in-batch indexes. The red stripes denote the conditional distribution of $y$ on a fixed $x$. PCL positive examples form `U' shapes and are highly correlated, their correlation constantly varying across batches. In contrast, GCL positive examples are more independent and obey a consistent distribution across batches.}
  \label{fig:distance_corr}
  \vspace{-0.4cm}
\end{figure}

\paragraph{Theoretical correlation in PCL positive distribution.} Based on our observation, PCL positive correspondences are clustered on roadside lines with a strong density correlation.
The equation describing a line $h$ meters away from a pair of vehicles that are $b$ meters apart follows equation \ref{eq:model}, where $x,y$ denotes the distance from a correspondence $c\in C$ to both LiDARs $L_a, L_b$, respectively. When using $x,y$ as the coordinates, the roadside line forms a `U' shaped pattern as depicted in Figure \ref{fig:hypothesis}. Our observation is also supported by Figure \ref{fig:corr_distribution}.

\begin{equation}
    ||\sqrt{x^2-h^2}\pm \sqrt{y^2-h^2}|| =b
    \label{eq:model}
\end{equation}

\vspace{-0.5cm}
\paragraph{Empirical correlation in PCL positive distribution.} We plot the real correspondences for both PCL (top row) and GCL (bottom row) in three different batches with $batch\_size=4$ in Figure \ref{fig:distance_corr}. Note that PCL (top row) exhibits several `U' shapes identical to Figure \ref{fig:hypothesis}(b). We conclude that the distribution of scan distance $y$ and $x$ are strongly correlated for PCL, violating the \textit{i.i.d.}~assumption.


\begin{figure}[t]
  \centering
  \includegraphics[width=0.7\linewidth]{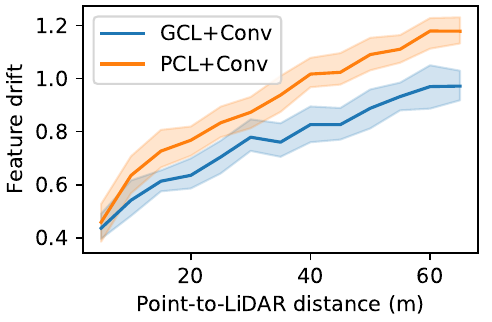}
  \vspace{-0.2cm}
  \caption{\textbf{Feature drift (as defined in Section \ref{sec:density_feat_corr}) grows with increasing distance from LiDAR.} The quality of GCL features deteriorates slower with decreasing density compared to PCL, indicating the superior density-invariance of GCL.}
  \label{fig:feature_drift}
  \vspace{-0.5cm}
\end{figure}

\vspace{-0.3cm}
\paragraph{Feature drift and its correlation with density.} Though long-desired in 3D vision, complete density invariance is unattainable due to its ill-posed nature, and deep features are generally affected by input point density. To quantitatively measure the degree of sensitivity to density-variation, we define a metric called \textit{feature drift} which represents the distance from a feature to the feature extracted from the densest observation of this spot. According to Figure \ref{fig:feature_drift}, there is a roughly linear correlation between the scan distance (quadratic-reciprocal with density) and the feature drift for convolutional backbones. The strong correlation between feature and density is the last piece of the puzzle. 
\label{sec:density_feat_corr}

\vspace{-0.3cm}
\paragraph{Summary of the analysis.}
Due to the density-mismatch in a pair of distant point clouds, PCL malfunctions under outdoor low-overlap scenario. Specifically, PCL samples correspondences in the limited overlap, resulting in a strong correlation between the point density in a correspondence. This undermines feature independence due to the linear relationship of density and feature. According to Lemma \ref{lemma:iid}, dependent positive features converge towards different locations so that they cannot be matched through Nearest-Neighbor Search, damaging registration performance.

Additionally, we believe that density correlation should also exist in other scenarios than self-driving. For example, a small indoor object will appear as clustered dots for PCL on Figure \ref{fig:distance_corr}, which is another kind of density correlation. Consequently, our analysis should still hold on other scenarios. Please refer to Appendix \ref{appendix:experiments} for more discussion.
\label{sec:sum_up}

\subsection{Overview of GCL}
\label{sec:overview}
\begin{figure}[t]
  \centering
  \includegraphics[width=1\linewidth]{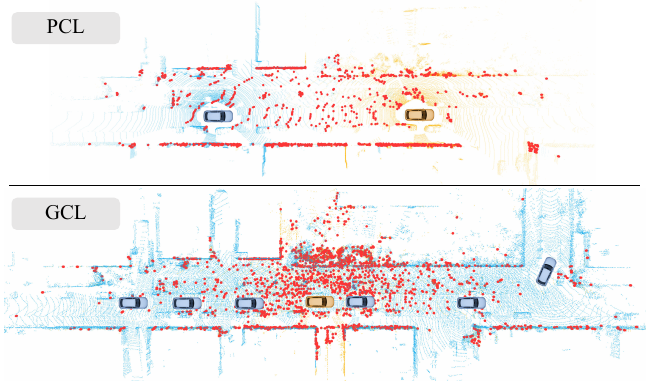}
  \caption{\textbf{Spatial distribution of 2000 random sampled positives denoted as red dots for PCL (top) and GCL (bottom).} $T$ and $C$ are tainted orange while $S$ and $\{N_k\}$ are tainted blue. PCL finds correspondences in the limited overlap, which severely biases towards roadside lines and ignores important object details both near and away from both LiDARs. In contrast, GCL positives are broadly scattered, obeying the same distribution as the central frame $C$.}
  \label{fig:corr_distribution}
  \vspace{-0.4cm}
\end{figure}

As depicted in Figure \ref{fig:overview}, during Point Cloud Sampling, GCL divides the region of $[-60m,60m]$ around a central frame $C$ into $\phi$ segments and select one point cloud from each of the segments with uniform distribution, forming the neighborhood point clouds $\{N_k|k=1,...,\phi\}$. Then, Positive Group Searching is carried out through nearest-neighbor search on the aligned point clouds to find multiple correspondences in the same spatial location, referred to as a positive group. As the point clouds complement each other to cover the whole space, positive groups could be found everywhere without special bias towards roadside lines, as depicted in Figure \ref{fig:corr_distribution}. The point clouds are then passed through a feature backbone to extract features for every point, and the corresponding features for positive groups are collected. Finally, group-wise losses are applied on positive groups to train the feature extraction network to be density-invariant.


\subsection{Positive Group Searching}
Intuitively, the whole central frame is the most consistent yet non-biased spatial distribution of points, which is ideal for positive sampling. We follow this idea to launch nearest-neighbor search from every single point in the central frame $C$ to all neighboring point clouds $\{N_k|k=1,...,\phi\}$, and gather all matches to form a positive group. Often having more than 2 matched points, the positive group can better approximate the underlying continuous high-dimensional feature distribution. It is still possible that some points in $C$ have no nearest neighbor and we simply discard them. Empirically, $87\%\pm5.4\%$ of the points in the central frame are able to form positive groups on KITTI.
Consequently, the density of GCL positives are far more independent and consistent than PCL, as depicted in Figure \ref{fig:distance_corr}.

\subsection{Group-wise Loss}
\label{sec:loss_design}
\begin{figure}[t]
  \centering
  \includegraphics[width=1\linewidth]{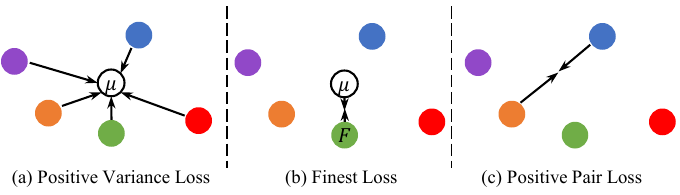}
  \caption{\textbf{Loss designs in positive groups.} Traditional Positive Pair Loss optimizes a pair of features, while GCL optimizes both the variance and the mean of a positive group through Positive Variance Loss and Finest Loss, respectively. $\mu$ denotes group mean while $F$ denotes finest feature (see Section \ref{sec:loss_design} for definition).}
  \label{fig:loss}
  \vspace{-0.3cm}
\end{figure}

In principle, the positive group loss should cater to positive groups instead of positive pairs to fully exploit in-group information. With a slight overload of subscript, we denote all positive groups as $G=\{g_x\}$ which are formulated as sets of $d$-dimensional features  $g_x=\{f_x^i\in \mathbb{R}^d\}$. Note that positive groups can be formulated using either points or features, as every point has its corresponding feature vector. All loss formulations are visualized in Figure \ref{fig:loss}. We formulate the positive pair loss $L_{PP}$ as Equation \ref{eq:loss_PP}, where $f_x^i, f_x^j$ are a single pair of correspondence sampled from $g_x$. For all loss formulations, parameter $r\in[1,+\infty)$ represents vector norm and $m_1,m_2,m_3,m_4\in \mathbb{R}^+$ are 4 tolerance margins.

\vspace{-0.2cm}
\begin{equation}
    L_{PP}=\frac{1}{|G|}\sum_{g_x\in G}{\max(||f_x^i-f_x^j||_r-m_1, 0)}
    \label{eq:loss_PP}
\end{equation}
\vspace{-0.2cm}

In contrast, we explore two new loss functions, namely Positive Variance Loss $L_{PV}$ and Finest Loss $L_F$. Positive Variance Loss minimizes the in-group feature variance to reduce sampling instability, as shown in Equation \ref{eq:loss_PV}, where $\mu(\cdot)$ is the average operator. Finest Loss in turn improves the subtle in-group structure, asking the mean feature to be close to the \textit{finest feature} as formulated in Equation \ref{eq:loss_F}. Finest feature is defined as the feature extracted on the point cloud with highest density in this group, and the function $\mathcal{F}(\cdot)$ returns the finest feature of a group.

\vspace{-0.2cm}
\begin{equation}
    L_{PV}=\sum_{g_x\in G}{\sum_{f_x^i\in g_x}{\frac{\max(||f_x^i-\mu(g_x)||_r-m_2, 0)}{|G|\times |g|}}}
    \label{eq:loss_PV}
\end{equation}

\vspace{-0.2cm}
\begin{equation}
    L_{F}=\sum_{g_x\in G}{\frac{\max(||\mathcal{F}(g_x)-\mu(g_x)||_r-m_3, 0)}{|G|}}
    \label{eq:loss_F}
\end{equation}
\vspace{-0.2cm}

We adopt the hardest-negative loss \cite{choy2019fully}, which is generalized to group-wise form as Equation \ref{eq:loss_HN}, where $\mathcal{H}(\cdot, \cdot, \cdot)$ defined in Equation \ref{eq:HN} returns the distance from a feature to its hardest negative, the latter defined as the nearest neighbor of the feature among all non-correspondence features. The total loss is defined as $L=\lambda_1 L_{PV} + \lambda_2 L_{F} + \lambda_3 L_{HN}$, where $\lambda_1,\lambda_2,\lambda_3$ controls the ratio between loss terms.

\vspace{-0.5cm}
\begin{equation}
    L_{HN}=\sum_{g_x\in G}{\sum_{f_x^i\in g_x}{\frac{\max(m_4 - \mathcal{H}(f_x^i, g, G), 0)}{|G|\times |g|}}}
    \label{eq:loss_HN}
\end{equation}

\vspace{-0.5cm}
\begin{equation}
    \mathcal{H}(f_x^i, g^x, G)=\min_{g_x\neq g_y\in G, f_y^j\in g_y}{||f_x^i - f_y^j||_r}
    \label{eq:HN}
\end{equation}


\section{Results}

We test GCL design on two outdoor vehicle-mounted LiDAR registration datasets KITTI \cite{Geiger2012CVPR} and nuScenes \cite{Caesar_2020_CVPR} with comparison and ablation studies. We then compare GCL to a mass sampling baseline and showcase the density invariance of GCL.

\subsection{Experiment Setup}
\paragraph{Datasets.} We validate GCL design on two commonly used outdoor registration datasets KITTI \cite{Geiger2012CVPR} and nuScenes \cite{Caesar_2020_CVPR}. We sub-divide the PCL datasets with different registration difficulty, measured by the distance between two LiDARs, using $[b_1,b_2]$ to denote that the distance is uniformly sampled between $b_1$ and $b_2$ in meters. These PCL datasets are used both during training of previous methods and testing of all methods. As SpinNet does not open-source its dataset preparation code, we report the test results on its pretrained models. The GCL datasets are only used during training, and are created differently as discussed in Section \ref{sec:overview} but consists of similar size to PCL datasets. We also distill low-overlap datasets with $\leq30\%$ overlap, denoted as LoKITTI and LoNuScenes following the methodology of Predator \cite{huang2021predator}. We follow the general protocols \cite{choy2019fully,Caesar_2020_CVPR} to divide KITTI and nuScenes into train-val-test splits. Please refer to Appendix \ref{appendix:exp_setup} for details.

\paragraph{Training.} As previous methods often struggle to converge on distant LiDAR point cloud pairs, we start by training a baseline model on $[5,20]$ and fine-tune 4 additional models on $[5,30], [5,40], [5,50], [5,60]$, then report the best performance of all models. However, there is no pair-wise LiDAR distance difference for GCL, so only one GCL model is trained and tested on arbitrary distances.

\begin{table}[t]
  \centering
  \small
  \resizebox{\linewidth}{1.7cm}{
  \begin{tabular}{@{}l|c|ccccc@{}}
    \toprule
    Dataset  & mRR	&\textit{[5,10]}	&\textit{[10,20]} &\textit{[20,30]} &\textit{[30,40]} &\textit{[40,50]}\\
    \midrule
    FCGF \cite{choy2019fully}	     &55.2      &97.0 	&85.4 	&54.1 	&25.0 	&14.3\\
    Predator \cite{huang2021predator}	   &74.7    &\underline{99.3}	&\bf{96.8}	&\underline{90.2}	&60.6	&26.7\\
    SpinNet \cite{ao2021spinnet}	      &35.6      &97.6	&73.1	&7.3	&0.0	&0.0\\
    D3Feat \cite{bai2020d3feat}	       &52.5     &98.7	&86.8	&52.7	&20.0   &4.5\\
    CoFiNet \cite{yu2021cofinet}	       &68.6    &\bf{99.6} 	&94.2 	&80.0 	&44.8 	&24.3\\
    GeoTransformer \cite{qin2022geometric}	&39.0    &97.9 	&88.3 	&8.3 	&0.7 	&0.0\\
    \midrule
    GCL+KPConv (ours)	    &\underline{83.5} &99.1 	&\underline{96.5} 	&89.3 	&\underline{78.6} 	&\underline{54.1}\\
    GCL+Conv (ours)     &\bf{88.8} &98.4 	&96.1 	&\bf{94.1} 	&\bf{87.6} 	&\bf{67.6}\\    \bottomrule
  \end{tabular}
  }
  \vspace{-0.2cm}
  \caption{\textbf{Comparison of RR (\%) between SOTA methods and GCL on five \textit{KITTI [$b_1,b_2$]} datasets}, with increasing LiDAR distance and registration difficulty. The mean RR is displayed in the first column.}
  \label{tab:comparison_kitti}
\end{table}

\begin{table}[t]
  \centering
  \small
  \resizebox{\linewidth}{1.05cm}{
  \begin{tabular}{@{}l|c|ccccc@{}}
    \toprule
    Dataset & mRR	&\textit{[5,10]}	&\textit{[10,20]} &\textit{[20,30]} &\textit{[30,40]} &\textit{[40,50]}\\
    \midrule
    FCGF \cite{choy2019fully} 	&37.0     &78.4 	&46.6 	&27.6 	&22.0 	&10.2 \\
    Predator \cite{huang2021predator} 	&39.5   &96.6	&50.9	&32.8	&9.8 	&7.4\\
    \midrule
    GCL-Conv (ours) 	&\underline{70.2} 	    &\underline{97.6} 	&\underline{88.0} 	&\underline{71.7} 	&\underline{56.7} 	&\bf{37.1}\\
    GCL-KPConv (ours)	&\bf{71.5} 	&\bf{99.0} 	&\bf{91.0} 	&\bf{77.7} 	&\bf{57.3} 	&\underline{32.5}\\
    \bottomrule
  \end{tabular}
  }
  \vspace{-0.2cm}
  \caption{\textbf{Comparison of RR (\%) between SOTA methods and GCL on five \textit{nuScenes [$b_1,b_2$]} datasets}, with increasing LiDAR distance and registration difficulty. The mean RR is displayed in the first column.}
  \vspace{-0.4cm}
  \label{tab:comparison_nuscenes}
\end{table}

\subsection{Performance Comparison}
We compare GCL against SOTA methods under five \emph{[$b_1$,$b_2$]} datasets on both \textit{KITTI} and \textit{nuScenes} with increasing registration difficulty, under a rigid registration criterion of $RTE\leq0.6m, RRE\leq1.5\degree$. The RR of all methods on KITTI and nuScenes are listed in Table \ref{tab:comparison_kitti} and Table \ref{tab:comparison_nuscenes}, respectively. The mean RR is shown in the first column. 

On \textit{KITTI} dataset, unfortunately, the training of D3Feat \cite{bai2020d3feat} and GeoTransformer \cite{qin2022geometric} do not converge, resulting in huge performance drop on harder datasets. Observing from the mRR, both GCL+Conv and GCL+KPConv are the best across all methods, achieving 88.8\% (+14.1\%) and 83.5\% (+8.8\%) mRR improvement over the closest competitor Predator. Despite performing on par with SOTA methods on the saturated easy datasets including \emph{KITTI [5,10]} and \emph{KITTI [10,20]}, GCL receives drastic improvements on scenarios with larger registration difficulty, achieving 67.6\% (+40.9\%) and 54.1\% (+27.4\%) RR on \emph{KITTI [40,50]} dataset compared to Predator, which proves the outstanding discriminative power and density invariance of GCL.

\begin{figure*}[t]
  \centering
  \includegraphics[width=1\linewidth]{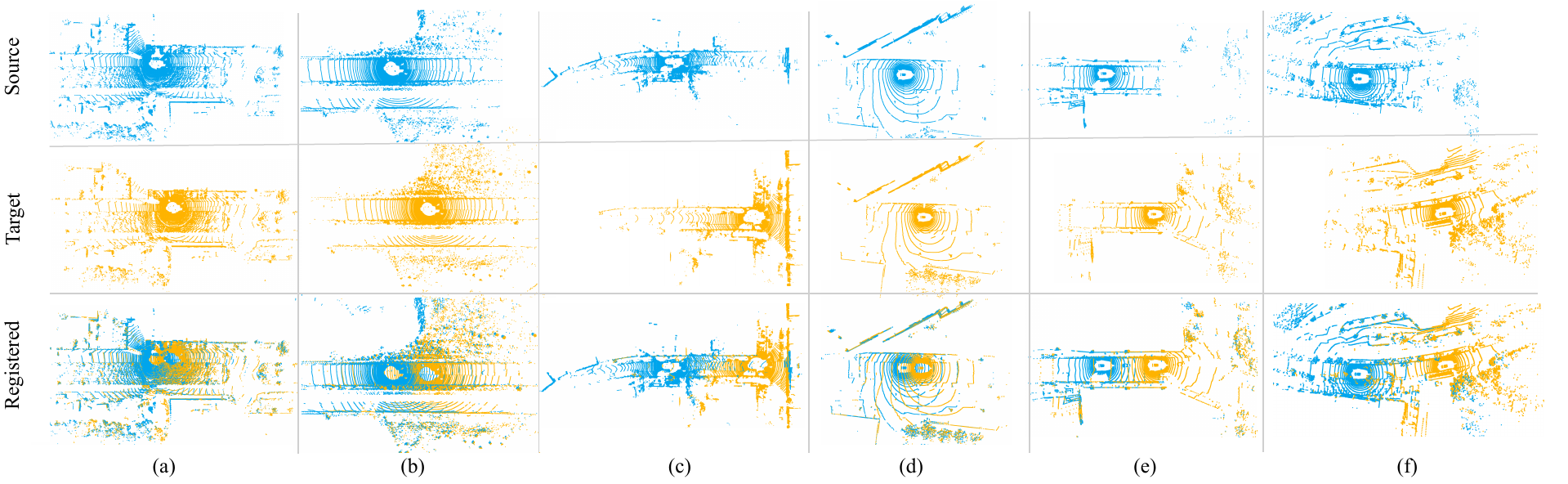}
  \vspace{-0.4cm}
  \caption{\textbf{Example registration results of GCL on \textit{KITTI} (a-c) and \textit{nuScenes} (d-f).} The point clouds are sampled from (a) \textit{KITTI [5,10]}, (b) \textit{KITTI [20,30]}, (c) \textit{KITTI [40,50]}, (d) \textit{nuScenes [5,10]}, (e) \textit{nuScenes [20,30]}, and (f) \textit{nuScenes [40,50]} datasets, respectively. Distant point clouds are both significantly harder to register and more rewarding to downstream tasks.}
  \vspace{-0.4cm}
  \label{fig:visualization}
\end{figure*}

On \textit{nuScenes} dataset, the improvements of GCL is much more broadly aware than those on \textit{KITTI}, achieving 70.2\% (+30.7\%) and 71.5\% (+32.0\%) mRR with GCL+Conv and GCL+KPConv, respectively. Dramatic improvements are visible even in close-range datasets including \textit{nuScenes [10,20]}, where 88.0\% (+37.1\%) and 91.0\% (+40.1\%) RR improvements are made by GCL+Conv and GCl+KPConv, respectively. While GCL+KPConv can deal with slight density variation on \textit{nuScenes [5,10]} to \textit{[30,40]}, GCL+Conv is better under extreme density variation as it performs the best on \textit{[40,50]} dataset both on KITTI and nuScenes. We conclude that the representative capability of GCL is even more significant on harder datasets such as nuScenes. Example registration results are placed in Figure \ref{fig:visualization}.

\subsection{Ablation Study}
\begin{table}[t]
  \centering
  \small
  \begin{tabular}{@{}lccc|ccc@{}}
    \toprule
    \multirow{2}{*}{Loss} & \multicolumn{3}{c|}{\cellcolor{lightgray}\emph{LoKITTI}}  & \multicolumn{3}{c}{\emph{KITTI [10,10]}}\\
    &\cellcolor{lightgray}RR &RTE &RRE  &RR &RTE &RRE\\
    &\cellcolor{lightgray}&\multicolumn{5}{c}{}\\
    \vspace{-0.76cm}\\
    \midrule
    \vspace{-0.41cm}\\
    C	 	&\cellcolor{lightgray}70.1	&27.0	&\underline{1.07}    &\underline{99.0}	&7.25	&0.28\\
    F	 	&\cellcolor{lightgray}53.2 	&37.1 	&1.43    &\bf{99.2}	&6.86 	&\underline{0.26}\\
    PP	    &\cellcolor{lightgray}70.3	&\underline{25.9}	&1.06   &98.6	&6.90	&0.28 \\
    PV	    &\cellcolor{lightgray}64.4	&\bf{24.2}	&\underline{1.07}	    &\bf{99.2}	&\bf{6.38}	&\textbf{0.24} \\
    BF+PP	&\cellcolor{lightgray}50.1	&34.7	&1.43 	&\bf{99.2}	&7.40	&0.31\\
    F+PP	&\cellcolor{lightgray}\underline{72.1}	&27.3	&0.97 	&\bf{99.2}	&7.11	&\underline{0.26}\\
    \textbf{F+PV}   &\cellcolor{lightgray}\bf{72.3}	&25.9	&\bf{1.03} 	&98.6	&\underline{6.62}	&\underline{0.26}	\\
    \bottomrule
  \end{tabular}
  \vspace{-0.2cm}
  \caption{\textbf{Ablation of loss designs} for GCL on \emph{KITTI [10,10]} and \emph{LoKITTI}, measured by RR (\%), RTE (cm), and RRE (\degree). F+PV is selected due to having the best performance on \textit{LoKITTI}. The gray column is the main metric.}
  \label{tab:loss_kitti}
  \vspace{-0.3cm}
\end{table}


\begin{table}[t]
  \centering
  \small
  \resizebox{\linewidth}{1.6cm}{
  \begin{tabular}{@{}lcc|cc@{}}
    \toprule
    \multirow{2}{*}{$\phi$} & \cellcolor{lightgray}RR (\%) on& RR (\%) on& Training  & Feature extraction\\
    &\cellcolor{lightgray}\textit{LoKITTI}	&\textit{KITTI [10,10]}	&time (h)	& time (ms)\\
    &\cellcolor{lightgray}&\multicolumn{3}{c}{}\\
    \vspace{-0.76cm}\\
    \midrule
    \vspace{-0.41cm}\\
    2 & \cellcolor{lightgray}46.3 & 98.6 & 32.1 & 44.2\\
    4 & \cellcolor{lightgray}68.8 & \textbf{98.8} & 38.6 & 44.2\\
    \textbf{6} & \cellcolor{lightgray}\textbf{72.3} & 98.6 & 58.6 & 44.2\\
    8 & \cellcolor{lightgray}71.4 & 98.6 & 71.0 & 44.2\\
    10 & \cellcolor{lightgray}69.7 & 98.4 & 78.1 & 44.2\\
    \bottomrule
  \end{tabular}
  }
  \vspace{-0.2cm}
  \caption{\textbf{Ablation on number of point clouds $\phi$} from 2 to 10 for GCL on \emph{KITTI [10,10]} and \emph{LoKITTI}. $\phi=6$ is selected to achieve the highest RR on \textit{LoKITTI}. The gray column is the main metric.
  }
  \vspace{-0.4cm}
  \label{tab:phi_kitti}
\end{table}

\paragraph{Loss ablation.} We ablate various GCL loss components and display the registration performance of GCL+Conv on both \textit{KITTI [10,10]} and \textit{LoKITTI} in Table \ref{tab:loss_kitti}. $C$ denotes Circle Loss \cite{sun2020circle} reimplemented on positive groups; $PP,F,PV$ are Positive Pair Loss, Finest Loss, and Positive Variance Loss, respectively; $BF$ is a variant of Finest Loss where the gradient of the finest feature is blocked, under the assumption that the finest feature should not be moved. Generally, most configurations perform similarly on \textit{KITTI [10,10]}. While F and PV are inferior to PP on their own, they perform better when combined. F+PP and F+PV performs the best on \textit{LoKITTI}, achieving 72.1\% and 72.3\% RR, respectively. We keep F+PV as the optimal loss configuration during other experiments.

\vspace{-0.3cm}

\paragraph{Number of point clouds $\phi$.} The number of point clouds is the major difference between GCL and PCL, as GCL with $\phi=1$ degrades to PCL. We ablate $\phi$ from 2 to 10 and list GCL performance and training time on \textit{KITTI [10,10]} and \textit{LoKITTI} in Table \ref{tab:phi_kitti}. While GCL with larger $\phi$ receives extended training time, the feature extraction time during inference stay constant. Considering the major metric (\emph{i.e.},RR on \textit{LoKITTI}), GCL performance quickly degrades with low $\phi$, while peaks at $\phi=6$. With $\phi>6$, the gain from GCL marginalizes, but the motional blur worsens with multiple frames, causing marginal degradation. We pick $\phi=6$ to achieve the best performance.

\begin{table}[t]
  \centering
  \small
  \vspace{0.2cm}
  \resizebox{\linewidth}{0.8cm}{
  \begin{tabular}{@{}lcccccc|c@{}}
    \toprule
    Data upscale & 1$\times$ & 2$\times$ & 3$\times$ &5$\times$ &10$\times$ &20$\times$ &GCL \\
    \midrule
    RR (\%) on \textit{LoKITTI} & 24.0 & 24.5 &25.2 &25.8 &27.9 &28.8 &72.3\\
    Time per epoch (min) & 6 & 12 & 19 & 34 & 59 &127 & 58\\
    \bottomrule
  \end{tabular}
  }
  \caption{\textbf{Comparison to a mass sampling baseline.} Scaling up the number of training point clouds of FCGF by up to 20$\times$ the original amount results in marginal improvements. Performance of GCL+Conv is displayed in the last column.
  }
  \label{tab:mass_sampling}
\end{table}

\vspace{-0.3cm}
\paragraph{Comparison to a mass sampling baseline.} A common misconception is that GCL succeeds simply through sampling several times more point clouds than PCL. In this study, we craft a baseline by increasing the number of sampled point cloud pairs for FCGF from 1$\times$ to 20$\times$. In particular, PCL with a $3.5\times$ upscale uses an equivalent number of point clouds as GCL with $\phi=6$. The performance of FCGF on \textit{LoKITTI} is displayed in Table \ref{tab:mass_sampling}, which indicates that mass sampling will only contribute marginally to network performance.

\begin{figure}[t]
  \centering
  \includegraphics[width=0.8\linewidth]{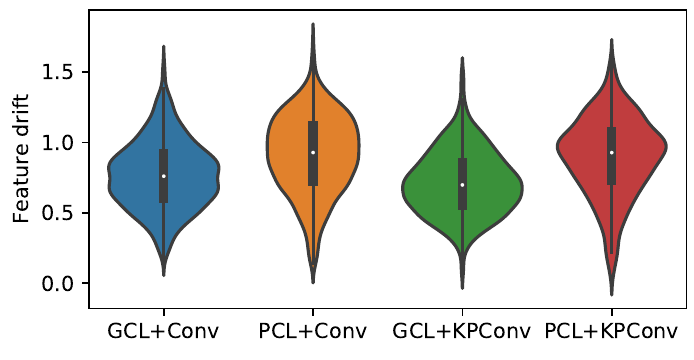}
  \vspace{-0.4cm}
  \caption{\textbf{Distribution of feature drift} (see Section \ref{sec:density_feat_corr} for definition) of GCL and PCL with sparse convolution and KPConv. GCL has constantly lower feature drift on two different convolutional architectures, which indicates the superior density-invariance and wide applicability of GCL.
  }
  \label{fig:density_invariance_violin}
  \vspace{-0.4cm}
\end{figure}

\vspace{-0.4cm}
\paragraph{Density Invariance.} 
We visualize the distribution of feature drift for PCL and GCL with two types of backbones in Figure \ref{fig:feature_drift} and Figure \ref{fig:density_invariance_violin}. Feature drift reveals the backbone's sensitivity to density change and should ideally stay low for complete density invariance.
GCL methods have consistently lower feature drift compared to the baselines, with both a lower average and smaller variance. This indicates that GCL-trained FCNs are more density-invariant than PCL-trained models.
\label{sec:density_invariance}


\vspace{-0.2cm}

\section{Conclusion}
\vspace{-0.1cm}

We have proposed GCL, a density-invariant feature extraction scheme for distant outdoor point cloud registration. Through the joint sampling and optimization on a group of point clouds, GCL naturally avoids being affected by the low-overlap and density-mismatch problem on a pair of distant point clouds. Fully convolutional networks trained with GCL are more representative and density-invariant than SOTA methods trained with PCL, thus being preferable for distant point cloud registration. GCL performs on par with SOTA methods on close point cloud pairs, but exhibits a drastic performance increase on distant point cloud pairs, resulting in a much higher overall performance. Specifically, GCL surpasses SOTA methods with +40.9\% RR on KITTI and +26.9\% RR on nuScenes in distant scenarios.

\vspace{-0.45cm}
\paragraph{Acknowledgement.} 
This work was supported in part by the National Natural Science Foundation of China (Grant No.\ 61972081), the Natural Science Foundation of Shanghai (Grant No.\ 22ZR1400200), and the Fundamental Research Funds for the Central Universities (No.\ 2232023Y-01). We would also like to thank reviewers and Li Chen from OpenDriveLab for fruitful discussions. 

{\small
\bibliographystyle{ieee_fullname}
\bibliography{egbib}
}

\clearpage
\appendix

\section{Proof of Lemma \ref{lemma:iid}}
\label{proof:iid}
Lemma \ref{lemma:iid} serves as the basis of our analysis, indicating the fundamental incompetence for PCL, while hinting an \textit{i.i.d.}~solution towards density-invariance.\\

\noindent \emph{Proof.} Since we need to investigate the effect of loss on a single feature $f^S_i$, we need to marginalize the effect of $f^T_j$. We start by selecting a specific first feature $\hat{f}^S_i$, regarding it as a constant, and take out all correspondences for the specific feature, which is $\hat{C}=\{(\hat{f}^S_i, f^T_j)\in C\}$. We focus on a part of the loss involving this specific $\hat{f}^S_i$, referred to as a function $\hat{L}_{pos}(\hat{f}^S_i)$, in equation \ref{eq:pos_func}.

\begin{equation}
    \hat{L}_{pos}(\hat{f}^S_i)=\frac{1}{|\hat{C}|}\sum_{(\hat{f}^S_i, f^T_j)\in \hat{C}}{max(||\hat{f}^S_i-f^T_j||_p-m, 0)}
    \label{eq:pos_func}
\end{equation}

Next, we marginalize the effect of $f^T_j$ through sampling infinitely many $f^T_j$. Assuming $\hat{f}^S_i, f^T_j$ are i.i.d., then countless $f^T_j$ approximates the distribution $D$. We can write out the limitation of $\hat{L}_{pos}$ when $|\hat{C}| \to \infty$ as Equation \ref{eq:pos_lim}.

\begin{equation}
    \lim_{|\hat{C}|\to \infty}{\hat{L}_{pos}}(\hat{f}^S_i)=\mathbb{E}_{f^T_j \sim D}{max(||\hat{f}^S_i, f^T_j||_p-m, 0)}
    \label{eq:pos_lim}
\end{equation}

Equation \ref{eq:pos_lim} is convex and has a single global minimum at $\hat{f}$ which solely depends on $D$ (cases where a minimal plateau exists is impossible in real setup). The effect of minimizing $L_{pos}$ converges in probability to all features $f\sim D$ heading towards the same location $\hat{f}$ in feature space.

\begin{equation}
     \lim_{|\hat{C}|\to \infty}{\hat{L}_{pos}}(\hat{f}^S_i)=\mathbb{E}_{(\hat{f}^S_i, f^T_j)\in \hat{C}}{max(||\hat{f}^S_i, f^T_j||_p-m, 0)}
    \label{eq:pos_lim_noniid}
\end{equation}

Otherwise, if $\hat{f}^S_i, f^T_j$ are non-\textit{i.i.d.}, it is impossible to marginalize $f^T_j$, and the loss in Equation \ref{eq:pos_lim_noniid} is the expectation on a subset of correspondences $\hat{C}$ that is correlated with $\hat{f}^S_i$. All likely features have different loss formulation with different global minimums. This means that different features will converge towards different locations.

Note that the loss we investigate is a partial representation of the complete loss function, as negative losses are not considered. However, the result is highly likely true even with negative loss added. That is because positive loss controls the sub-structure inside a specific positive cluster, while negative loss controls the large-scale relative structure between different positive clusters, and the negative loss should not disturb positive structures too much when the feature representation stabilizes after the first few epochs.

\section{Detailed Experiment Setup}
\label{appendix:exp_setup}

\subsection{Dataset Preparation}
Two kinds of datasets are used in this paper, \emph{i.e.}, pair-wise contrastive learning (PCL) datasets and group-wise contrastive learning (GCL) datasets. The PCL datasets contain point cloud pairs that are sampled with a random distance interval $b$ denoting the distance between two LiDARs. The distance $b$ is randomly picked for every point cloud pair, and we refer to a sub-divided dataset where $b_1\leq b \leq b_2$ as \textit{[$b_1$,$b_2$]}. Both during training and testing, we always reset the random seed to 0 before finding the required point clouds to produce the exact same point cloud pairs for repeatable results. To create the GCL datasets, we sample central point clouds $C$ at a fixed interval of 11 frames, then randomly sample neighboring point clouds around each central point cloud according to the process described in Section \ref{sec:overview}. The GCL datasets are never used during testing.

Following Huang \etal \cite{huang2021predator}, we define \textit{overlap} $O$ between a pair of point clouds $S\in\mathbb{R}^{N\times 3},T\in\mathbb{R}^{M\times 3}$ as a subset of $S$ according to Equation \ref{eq:overlap}.

\begin{equation}
    O=\left\{p_S^i\in S\ \Bigg|\ \min_{P_T^j\in T}\|p_S^i-p_T^j\|_2 \leq \delta\right\}
    \label{eq:overlap}
\end{equation}

The overlap denotes the part of $S$ where at least a corresponding point in $T$ could be found through nearest-neighbor search of radius $\delta=0.45m$. $S$ and $T$ are down-sampled using a voxel size of 0.3m before the search.  Overlap ratio is then defined as $\frac{|O|}{|S|}$. All point cloud pairs with $\leq 30\%$ overlap ratio in \textit{[5,20], [20,30], [30,40], [40,50]} datasets are collected on \textit{KITTI} and \textit{nuScenes}, referred to as \textit{LoKITTI} and \textit{LoNuScenes}, respectively. They represent the hardest cases for the distant point cloud registration task. 

We follow previous literature \cite{bai2020d3feat} to divide OdometryKITTI with sequences 0-5 for training, 6-7 for validation, and 8-10 for testing. NuScenes is divided sequentially with the first 700 sequences for training, the next 150 sequences for validation and the last 150 sequences for testing.


\subsection{Metrics}
Both traditional and new metrics are used during evaluation. Following previous work \cite{gojcic2018learned,huang2021predator,bai2020d3feat,choy2019fully}, we report 3 metrics including Registration Recall (RR) defined as percentage of pairs successfully registered, Relative Rotation Error (RRE) defined as the geodesic distance between estimated rotation and ground-truth rotation, and Relative Translation Error (RTE) defined as the euclidean distance between estimated translation and the ground-truth translation. We forge a new metric as the average of RR on $[5,10], [10,20], [20,30], [30,40], [40,50]$ datasets, referred to as mean Registration Recall (mRR), which measures the overall registration performance.


\subsection{Network Structure}
\begin{figure*}[t]
  \centering
  \includegraphics[width=1\linewidth]{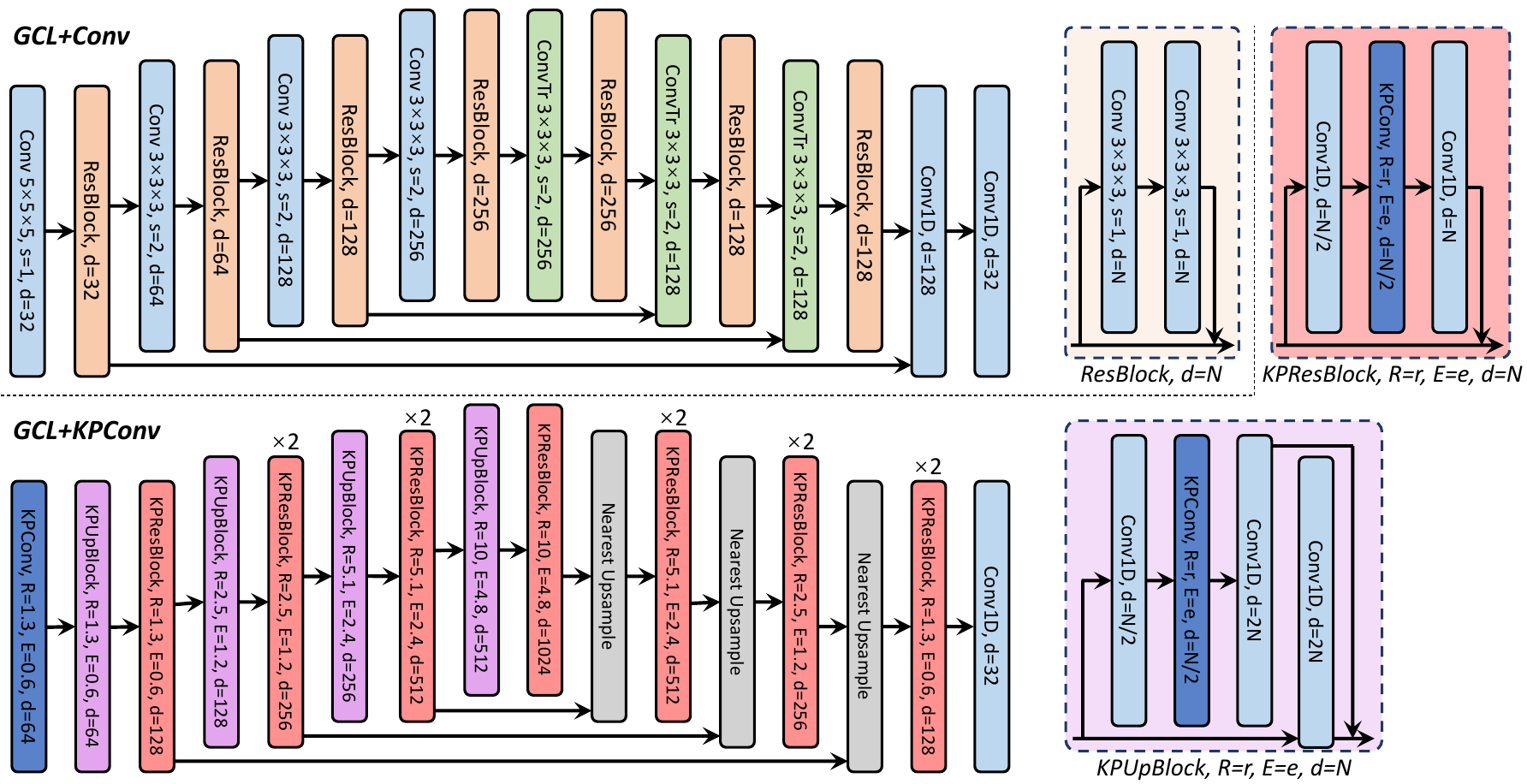}
  \vspace{-0.4cm}
  \caption{\textbf{Network structures for GCL.} Batch Normalization and ReLU activation are used after all Conv blocks except for the last layer, while batch normalization and leaky ReLU are used in KPConv blocks with a 0.1 slope. Voxel Convolution is parameterized by the kernel size, stride $s$, and output dimension $d$. The kernel size and stride are both omitted for Conv1D. Non-deformable KPConv \cite{thomas2019kpconv} is parameterized by kernel point offset radius $R$, kernel point influence extent $E$, and feature dimension $d$.}
  \label{fig:network_structure}
  \vspace{-0.4cm}
\end{figure*}

We adopt the popular Res-UNet network structure~\cite{choy2019fully}, and implement it on both sparse voxel convolution~\cite{choy20194d} and KPConv~\cite{thomas2019kpconv}, referred to as GCL+Conv and GCL+KPConv, respectively. As depicted in Figure \ref{fig:network_structure}, both GCL+Conv and GCL+KPConv adopt three layers of skip connections with a roughly symmetric encoder-decoder design. Features are all normalized onto a unit sphere after the final layer.

\subsection{Loss Configuration}
There are several parameters that need specifying for network convergence. The distance margins are set to $m_1=0.1, m_2=0.1, m_3=0.2, m_4=1.4$. The loss terms are reweighed differently on two datasets, where we set $\lambda_1=\lambda_2=\lambda_3=1$ on KITTI and $\lambda_1=\lambda2=0.7, \lambda_3=1$ on nuScenes.

\section{Additional Experiments}
\label{appendix:experiments}

\begin{table}[t]
  \centering
  \small
  \resizebox{\linewidth}{1.2cm}{
  \begin{tabular}{@{}lccc|c@{}}
    \toprule
        	&Dataloader	&Inference &RANSAC &Total\\
    \midrule
FCGF	    &6.2	&45.4	&576.1	&627.7\\
GCL+Conv (ours)	&5.4	&44.2	&523.0	&572.6\\
\midrule
Predator	&635.1	&78.7	&66.3	&780.1\\
GCL+KPConv (ours)	&637.5	&64.4	&76.4	&778.3\\
    \bottomrule
  \end{tabular}
  }
  \caption{\textbf{Inference time (ms) analysis on \textit{LoKITTI}.} GCL is always more lightweight than their existing counterparts with the same backbone (FCGF: Conv; Predator: KPConv) in terms of inference time.
  }
  \label{tab:inference}
\end{table}
\paragraph{Inference time.} We list the inference time breakdown for FCGF, Predator, GCL+Conv and GCL+KPConv in Table \ref{tab:inference}. The inference time of GCL is always lower than counterparts with the same backbone. While GCL+KPConv performs faster registration, it requires extended data loading time due to underlying KPConv architecture conducting repeated nearest neighbor calculation. In contrast, GCL+Conv runs faster during data loading and inference, and the extended RANSAC registration time can be reduced given recent progress on fast registration pipelines \cite{bai2021pointdsc}. The focus of GCL is to propose a contrastive learning based training method which can be plugged into any existing registration pipelines that incorporate feature matching in it ~\cite{deng2018ppfnet,gojcic2019perfect,poiesi2021distinctive,ao2021spinnet,yew20183dfeat,choy2019fully,bai2020d3feat,huang2021predator,yew2022regtr}, and GCL is the general solution to the distant registration problem on all these methods since they are all based on either Voxel Convolution \cite{choy20194d} or KPConv \cite{thomas2019kpconv}. We conclude that GCL is a universal lightweight feature extraction method.

\begin{table}[t]
  \centering
  \small
  \begin{tabular}{@{}lccc|ccc@{}}
    \toprule
    \multirow{2}{*}{Loss} & \multicolumn{3}{c|}{\cellcolor{lightgray}\emph{LoKITTI}}  & \multicolumn{3}{c}{\emph{KITTI [10,10]}}\\
    &\cellcolor{lightgray}RR &RTE &RRE  &RR &RTE &RRE\\
    &\cellcolor{lightgray}&\multicolumn{5}{c}{}\\
    \vspace{-0.76cm}\\
    \midrule
    \vspace{-0.41cm}\\
C	    &\cellcolor{lightgray}\underline{53.8} 	&32.5 	&1.41 &\underline{99.0} 	&7.8 	&0.27 \\
F    	&\cellcolor{lightgray}18.3 	&38.9 	&1.92 	    &98.8 	&\underline{7.6} 	&\textbf{0.25}\\
PP	   	&\cellcolor{lightgray}\underline{53.8}	&\textbf{27.2}	&\textbf{1.28} &\textbf{99.2}	&\underline{7.6}	&\underline{0.26}\\
PV	    &\cellcolor{lightgray}45.0 	&29.1 	&1.39 &98.6 	&\underline{7.6} 	&\textbf{0.25}\\
BF+PP	&\cellcolor{lightgray}45.7 	&31.1 	&1.40 &98.6 	&\underline{7.6} 	&\underline{0.26}\\
F+PV	&\cellcolor{lightgray}50.5 	&28.4 	&\underline{1.30} &\textbf{99.2} 	&\textbf{7.5} 	&\textbf{0.25}\\
\textbf{F+PP}	&\cellcolor{lightgray}\textbf{55.4} 	&\underline{27.8} 	&\textbf{1.28 } &\textbf{99.2} 	&7.9 	&\underline{0.26}\\
    \bottomrule
  \end{tabular}
  \vspace{-0.2cm}
  \caption{\textbf{Ablation of loss designs} for GCL+KPConv on \emph{KITTI [10,10]} and \emph{LoKITTI}, measured by RR (\%), RTE (cm), and RRE (\degree). F+PP is selected according to performance on \textit{LoKITTI}. The gray column is the main metric.}
  \label{tab:loss_kitti_kpconv}
  \vspace{-0.4cm}
\end{table}

\paragraph{Loss ablation with GCL+KPConv.}  We ablate various loss components for GCL+KPConv and display the registration performance of on both \textit{KITTI [10,10]} and \textit{LoKITTI} in Table \ref{tab:loss_kitti_kpconv}. Similar to results with GCL+Conv, Finest Loss in combination with a positive loss performs the best among all methods, as F+PP achieves both the best RR of 55.4\% on \textit{LoKITTI} and 99.2\% \textit{KITTI [10,10]}. All methods perform roughly the same on the close point cloud dataset\textit{ KITTI [10,10]}. With the KPConv backbone, however, Finest loss alone does not lead to a decent performance on LoKITTI as it does with the voxel convolution backbone. We select F+PP as the optimal configuration for GCL+KPConv during all other experiments.

\begin{table}[t]
  \centering
  \small
  \resizebox{\linewidth}{1.7cm}{
  \begin{tabular}{@{}l|c|ccccc@{}}
    \toprule
    Dataset  & mRR	&\textit{[5,10]}	&\textit{[10,20]} &\textit{[20,30]} &\textit{[30,40]} &\textit{[40,50]}\\
    \midrule
    FCGF \cite{choy2019fully}               &77.4 	&98.4 	&95.3 	&86.8 	&69.7 	&36.9 \\
    Predator \cite{huang2021predator}       &87.9 	&\textbf{100.0}   &98.6	&\textbf{97.1}	&80.6	&63.1 \\
    SpinNet \cite{ao2021spinnet}            &39.1 	&99.1 	&82.5 	&13.7 	&0.0 	&0.0 \\
    D3Feat \cite{bai2020d3feat}             &66.4 	&99.8 	&98.2 	&90.7 	&38.6 	&4.5 \\
    CoFiNet \cite{yu2021cofinet}            &82.1 	&\underline{99.9} 	&\textbf{99.1} 	&94.1 	&78.6 	&38.7 \\
    GeoTransformer \cite{qin2022geometric}  &42.2 	&\textbf{100.0} 	&93.9 	&16.6 	&0.7 	&0.0 \\
    \midrule
    GCL+KPConv (ours)                       &\underline{89.6} 	&\textbf{100.0} 	&98.2 	&93.2 	&\underline{88.3} 	&\underline{68.5} \\
    GCL+Conv (ours)                         &\textbf{93.5} 	&99.0 	&\underline{98.8} 	&\underline{96.1} 	&\textbf{91.7} 	&\textbf{82.0} \\
    \bottomrule
  \end{tabular}
  }
  \caption{\textbf{Comparison of RR (\%) between SOTA methods and GCL on five \textit{KITTI [$b_1,b_2$]} datasets}, with increasing LiDAR distance and registration difficulty. Registration metrics are loosened to 5\degree, 2m compared to Table \ref{tab:comparison_kitti}. The mean RR is displayed in the first column.}
  \label{tab:comparison_kitti_loose}
\end{table}

\begin{table}[t]
  \centering
  \small
  \resizebox{\linewidth}{1.05cm}{
  \begin{tabular}{@{}l|c|ccccc@{}}
    \toprule
    Dataset & mRR	&\textit{[5,10]}	&\textit{[10,20]} &\textit{[20,30]} &\textit{[30,40]} &\textit{[40,50]}\\
    \midrule
    FCGF \cite{choy2019fully}           &39.5 	&87.9 	&63.9 	&23.6 	&11.8 	&10.2\\ 
    Predator \cite{huang2021predator}   &51.0 	&\underline{99.7}	&72.2	&52.8	&16.2	&14.3\\
    \midrule
    GCL-Conv (ours)                     &\underline{85.5} 	&99.3 	&\underline{97.7} 	&\underline{91.8} 	&\underline{77.8} 	&\underline{60.7}\\ 
    GCL-KPConv (ours)                   &\textbf{90.3}	&\textbf{99.9} 	&\textbf{98.5} 	&\textbf{96.1} 	&\textbf{85.4} 	&\textbf{71.6}\\ 
    \bottomrule
  \end{tabular}
  }
  \caption{\textbf{Comparison of RR (\%) between SOTA methods and GCL on five \textit{nuScenes [$b_1,b_2$]} datasets}, with increasing LiDAR distance and registration difficulty. Registration metrics are loosened to 5\degree, 2m compared to Table \ref{tab:comparison_nuscenes}. The mean RR is displayed in the first column.}
  \label{tab:comparison_nuscenes_loose}
\end{table}

\paragraph{Performance comparison under loose registration criterion.} We additionally provide the comparison between GCL and SOTA methods on both \textit{KITTI} and \textit{nuScenes} under a loose registration criterion of $RTE\leq 2m, RRE\leq 5\degree$, where the registration recalls are generally elevated due to the loosen criterion. The mean RR is shown in the first column. As listed in Table \ref{tab:comparison_kitti_loose},  GCL+Conv and GCL+KPConv achieve the highest overall performance on \textit{KITTI} with 89.6\% (+1.7\%) and 93.5\% (+4.6\%) mRR over Predator \cite{huang2021predator}, respectively. Furthermore, GCL methods receive greater improvements on distant scenarios including \textit{[30,40]} and \textit{[40,50]} on \textit{KITTI}. On the other hand, GCL methods beat SOTA methods by a larger margin on \textit{nuScenes} than on \textit{KITTI}, achieving 85.5\% (+34.5\%) and 90.3\% (+39.3\%) mRR for GCL+Conv and GCL+KPConv compared to Predator \cite{huang2021predator}, respectively on \textit{nuScenes} according to Table \ref{tab:comparison_nuscenes_loose}. We mark that GCL methods beat SOTAs on every sub-divided dataset on \textit{nuScenes}, and that GCL+KPConv always performs the best. We conclude that, under a loose registration criterion, GCL still achieves giant improvements comparable to the scenario under a stricter criterion, setting a new SOTA for the distant point cloud registration problem.

\begin{figure}[t]
  \renewcommand{\thefigure}{11}
  \centering
  \includegraphics[width=\linewidth]{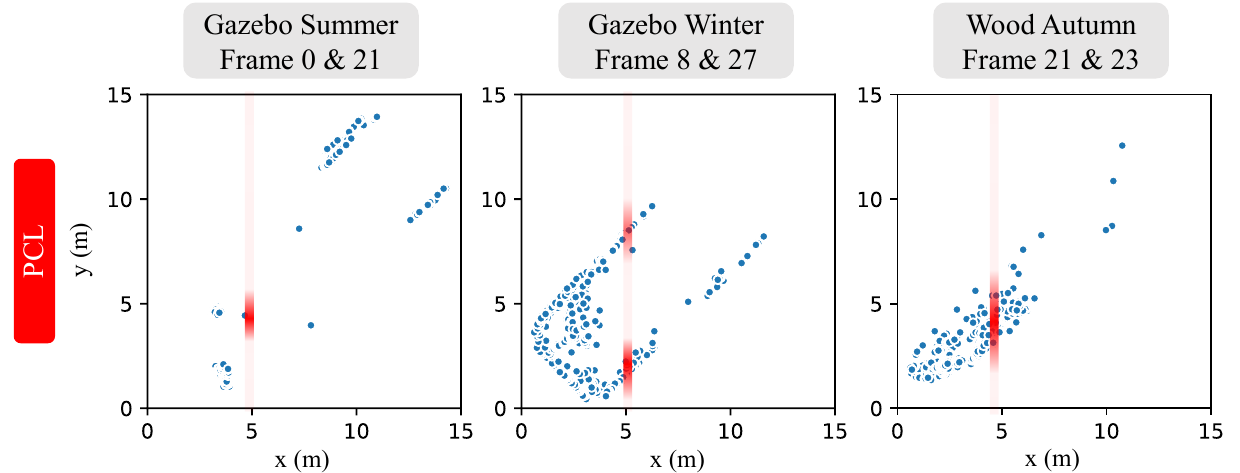}
  \caption{\textbf{Distribution of correspondences in ETH dataset} with PCL and $batch\_size=1$, where the $x$ and $y$ coordinates of a dot denotes the distance from a correspondence to both LiDARs. The red stripes denote the conditional distribution of $y$ on a fixed $x$. Density correlation still exists in ETH, where linear structures are less dominant.}
  \label{fig:eth}
\end{figure}

\begin{table}[t]
  \renewcommand{\thetable}{10}
  \centering
  \small
  \resizebox{\linewidth}{1.2cm}{
  \begin{tabular}{@{}l|cc|cc|c@{}}
    \toprule
    &&&&\vspace{-0.45cm}\\
    & \multicolumn{2}{c|}{\textbf{Gazebo}} 	& \multicolumn{2}{c|}{\textbf{Wood}} & \multirow{2}{*}{Avg.}\\
    & Summer & Winter & Autumn & Summer & \\
    &&&&\vspace{-0.4cm}\\
    \midrule
    Predator & 21.2 & 20.8 & 23.5 & 30.4 & 24.0 \\
    FCGF & 40.2 & 26.0 & 54.8 & 67.2 & 47.0\\
    GCL+KPConv (ours) &  46.2 & 28.4 & 56.5 & 72.0 & 50.8 \\
    GCL+Conv (ours) & \textbf{46.7} & \textbf{30.8} & \textbf{61.7} & \textbf{73.6} & \textbf{53.2}\\
    \multicolumn{6}{c}{\vspace{-0.45cm}}\\
    \bottomrule
  \end{tabular}
  }
  \caption{\textbf{Generalization test from KITTI to ETH}, by shrinking voxel size from 0.3m to 0.05m during testing. The FMR scores at $\tau_1$ = 10cm, $\tau_2$ = 5\% are compared.}
  \label{tab:generalize_eth}
\end{table}

\paragraph{Generalization to ETH.} We demonstrate the generalization results from KITTI to ETH in Table \ref{tab:generalize_eth}, by shrinking the voxel sizes from 0.3m to 0.05m during testing without any finetuning. ETH is an outdoor dataset featuring a majority of vegetation over linear structures. However, density correlation still exists in ETH, as depicted in Figure \ref{fig:eth}, which confirms the wide applicability of our analysis in Section \ref{sec:analysis}. It can be seen that GCL effectively improves the generalization capability of two baseline backbones, where GCL+Conv achieves the best overall FMR of 53.2\% (+6.2\%). We conclude that GCL can generalize to other scenarios other than autonomous driving.

\section{Discussion and Limitation}
\paragraph{More explanation on non-\textit{i.i.d.}~PCL positives.} A pair of close-range point clouds also have non-\textit{i.i.d.}~positives, as their positives have roughly the same density, \emph{i.e.}, their densities are positively correlated. This may sound weird, as close-range LiDAR point cloud registration has already been well-solved \cite{huang2021predator,yu2021cofinet}. Actually, non-\textit{i.i.d.}~positives will not hinder close-range registration problems because the problem is so simple that even a density-variant feature extractor will solve the problem nicely. Now consider a hand-crafted density-variant feature that upon the input coordinate $(x,y,z)$, outputs the vector length of the coordinate $\sqrt{x^2+y^2+z^2}$. Intuitively, this density-variant feature combined with RANSAC will likely produce a decent guess for two concentric (\emph{i.e.}, extremely close) point clouds. However, this special solution will not work for distant scenarios with severe density mismatch, which means that a more powerful solution like GCL is needed to solve the distant point cloud registration problem.
\paragraph{Training time.} As listed in Table \ref{tab:phi_kitti}, GCL has a linearly growing training time consumption \emph{w.r.t.}~$\phi$. This is mainly caused by increased data loading time where repeated nearest neighbor searches are carried out from the central point cloud to all neighborhood point clouds. However, the heavy time consumption is a necessary cost for building the positive groups. Luckily, only training time is affected for GCL and the testing time remain unchanged when registering two point clouds.
\paragraph{Information exchange.} Information exchange serves as a key source of improvement for SOTA registration methods \cite{huang2021predator,yu2021cofinet,wu2021feature,li2022lepard,qin2022geometric}. It is carried out between a pair of point clouds, which calls for a non-trivial extension of GCL that contains $\phi+1$ point clouds. Note that features after the exchange will vary according to different companion point clouds. 
Consequently, a naive traversal of $C_{\phi+1}^2$ pairs for GCL will not only suffer from $O(\phi^2)$ complexity but also have to deal with $\phi$ different features for a single point.
We hope to extend the information exchange module (mainly composed of cross-attention) to a group-wise version in future work.


\end{document}